\documentclass{article}
\usepackage{amssymb}

\usepackage{amsmath}
\usepackage{graphicx}
\usepackage{caption}
\usepackage{subcaption}
\usepackage{siunitx}
\usepackage{dsfont}
\usepackage{wrapfig}
\usepackage{paralist}
\usepackage{hyperref}

\usepackage[final]{corl_2025} 

\title{Constraint-Aware Diffusion Guidance for Robotics: Real-Time Obstacle Avoidance for Autonomous Racing}

\newcommand{\ETH}{Department of Mechanical and Process Engineering, ETH Z\"urich, Switzerland.}
\newcommand{\MPI}{Max Planck Institute for Intelligent Systems, T\"ubingen, Germany.}

\author{
  Hao Ma\thanks{\ETH}\\
  \texttt{haomah@ethz.ch} \\
  \And
  Sabrina Bodmer\footnotemark[1] \\
  \texttt{sabodmer@ethz.ch} \\
  \AND
  Andrea Carron\footnotemark[1] \\
  \texttt{carrona@ethz.ch} \\
  \And
  Melanie Zeilinger\footnotemark[1]~~\thanks{Shared last author.}\\
  \texttt{mzeilinger@ethz.ch} \\
  \And
  Michael Muehlebach\thanks{\MPI}~~\footnotemark[2]\\
  \texttt{michaelm@tuebingen.mpg.de}
}

\begin{document}
\maketitle


\begin{abstract}
    Diffusion models hold great potential in robotics due to their ability to capture complex, high-dimensional data distributions. However, their lack of constraint-awareness limits their deployment in safety-critical applications. We propose Constraint-Aware Diffusion Guidance (CoDiG), a data-efficient and general-purpose framework that integrates barrier functions into the denoising process, guiding diffusion sampling toward constraint-satisfying outputs. CoDiG enables constraint satisfaction even with limited training data and generalizes across tasks. We evaluate our framework in the challenging setting of miniature autonomous racing, where real-time obstacle avoidance is essential. Real-world experiments show that CoDiG generates safe outputs efficiently under dynamic conditions, highlighting its potential for broader robotic applications. A demonstration video is available at \href{https://youtu.be/KNYsTdtdxOU}{https://youtu.be/KNYsTdtdxOU}. 
\end{abstract}



\section{Introduction}
\label{sec:introduction}

Since their inception~\citep{sohl-dicksteinDeepUnsupervisedLearning,hoDenoisingDiffusionProbabilistic2020a}, diffusion models have achieved groundbreaking success in image~\citep{hoCascadedDiffusionModels,batzolisConditionalImageGeneration2021}, audio~\citep{jeongDiffTTSDenoisingDiffusion2021}, and video generation~\citep{hoVideoDiffusionModels2022}. Due to their exceptional capability in modeling multimodal data and capturing complex high-dimensional distributions, they have recently also garnered significant attention in robotics~\citep{wolfDiffusionModelsRobotic2025,chiDiffusionPolicyVisuomotor2024,urainSEDiffusionFieldsLearning2023}. Collectively, these works highlight how diffusion models address limitations of traditional policy approaches, such as unimodal assumptions or training instability, thereby offering a more versatile framework for robotic behavior learning.

While diffusion models hold significant promise for robotics, standard formulations still face important challenges related to safety and physical feasibility. Many approaches are trained purely on data without explicitly enforcing constraints, which can lead to collisions or dynamic infeasibility, particularly when encountering out-of-distribution scenarios~\citep{kondoCGDConstraintGuidedDiffusion2024,paloDiffusionAugmentedAgents2025}. Additionally, existing methods often rely heavily on large-scale offline datasets to promote generalization, which can limit their adaptability in unseen environments~\citep{leeLearningDiverseRobot2025,liangAdaptDiffuserDiffusionModels2023}. Addressing these challenges is crucial for enabling the safe and reliable deployment of diffusion models in safety-critical robotic applications.


To overcome these limitations, we propose Constraint-Aware Diffusion Guidance (CoDiG), a general-purpose, data-efficient diffusion-based framework for real-time, safe trajectory generation in robotics tasks such as autonomous racing. CoDiG integrates a barrier function directly into the reverse diffusion process, steering the sampling away from unsafe or dynamically infeasible regions without relying on external classifiers or auxiliary models. To further accelerate sampling and enhance its stability, CoDiG employs a warm-start strategy by initializing the diffusion process near feasible solutions. By augmenting the score updates with barrier gradients during inference, CoDiG enforces safety constraints, enabling efficient and reliable deployment in safety-critical environments. Our main contributions are summarized as follows:
\begin{itemize}
\item We introduce Constraint-Aware Diffusion Guidance (CoDiG), a general-purpose and data-efficient trajectory generation framework that enforces constraints during inference, allowing safe and physically feasible generalization from a small set of expert demonstrations to novel scenarios.
\item We propose a warm-start strategy that significantly accelerates the inference process, achieving real-time performance suitable for high-frequency, safety-critical control, while ensuring smooth transitions between trajectories generated at successive iterations.
\item We deploy CoDiG on a real-world autonomous racing car, tracking the trajectories generated by CoDiG and demonstrating safe obstacle avoidance in dynamic scenarios.
\end{itemize}

\section{Related Work}
\label{sec:related_work}
Many recent works have explored incorporating constraints into diffusion models for robotic tasks. Some approaches enforce constraints during training: \citet{bastekPhysicsInformedDiffusionModels2025} integrate physical laws into the training objective to ensure physically consistent outputs; \citet{giannoneAligningOptimizationTrajectories2023} align sampling trajectories with constrained optimization paths; and \citet{powerSamplingConstrainedTrajectories} separately train on different constraints and combine them at inference. Others address constraints during inference: \citet{carvalhoMotionPlanningDiffusion2023} condition the sampling process on goal-reaching and obstacle avoidance; \citet{christopherConstrainedSynthesisProjected2024,xiaoSafeDiffuserSafePlanning2023} enforce feasibility through projection steps, albeit with significant computational overhead; \citet{romerDiffusionPredictiveControl2024} incorporate model-based projections directly into the backward diffusion process to enforce constraint satisfaction during trajectory generation, avoiding the need for external post-sampling corrections; and \citet{yuLDPLocalDiffusion2024a} generate local collision-free motions through conditional sampling. Several methods handle constraints in both training and inference phases, such as \citet{ajayConditionalGenerativeModeling2023} for decision-making, \citet{OfflineSafeReinforcement} with trajectory-level diffusion, and \citet{botteghiTrajectoryGenerationControl2023}, which train safe priors and apply runtime filtering. Among these, \citet{yuLDPLocalDiffusion2024a} primarily handle inference-time constraints, while \citet{botteghiTrajectoryGenerationControl2023} combine both stages. Overall, incorporation during training time promotes inherent feasibility, while inference-time methods offer flexibility at the cost of higher computational complexity during inference.

Compared to prior inference-time approaches, our CoDiG framework handles constraints by augmenting score updates with lightweight barrier gradients during sampling, without relying on projections, auxiliary models, or simulators. This provides efficient, continuous guidance toward feasible trajectories while preserving the generative flexibility of diffusion models. Warm-start initialization further accelerates convergence and enhances sampling stability, enabling real-time deployment. Unlike previous works mainly evaluated in simulation or in quasi-static environments, we demonstrate CoDiG on a real-world autonomous racing platform, where strict dynamic feasibility and rapid obstacle avoidance are critical. While \citet{QuantizationFreeAutoregressiveAction} have also suggested autoregressive architectures as an alternative to diffusion-based generation, it is unclear whether constrained-aware generation via barrier functions is also effective with these architectures. These aspects highlight the unique contributions of CoDiG in enabling efficient, reliable, and real-time constraint handling within generative robotic planning.

\section{Methodology}
\label{sec:methodology}
Recent advances in score-based generative modeling formulate diffusion processes via stochastic differential equations (SDEs), offering a continuous-time view of forward noise injection and reverse denoising~\citep{songScoreBasedGenerativeModeling2021}. Since our work builds on this foundation, we briefly review score-based generative modeling and introduce the notation used throughout the paper.


\subsection{Preliminaries}
\label{sec:preliminaries}
Let $x_0 \in \mathbb{R}^d$ denote a noise-free data sample drawn from the data distribution $p_{0}(x)$. A score-based generative model defines a continuous-time diffusion process $\{x_t\}_{t \in [0,T]}$, where $t$ denotes the diffusion time, such that $x_T$ becomes approximately Gaussian. It is important to note that throughout this paper, we encounter two notions of ``time'': here, $t$ refers to the artificial diffusion time governing the processes, while later, $\tau$ will denote the physical time in real-world dynamical systems.

\paragraph{Diffusion Process.} The forward diffusion process gradually perturbs the data by solving the following SDE:
\begin{equation*}
    \mathrm{d}x_t = f(x_t, t)\,\mathrm{d}t + g(t)\,\mathrm{d}w_t, \quad t \in [0, T],\quad x_0\sim p_0,
\end{equation*}
where $x_t \in \mathbb{R}^d$ is the perturbed data at time $t$, $f: \mathbb{R}^d \times [0, T] \to \mathbb{R}^d$ is the drift term, $g: [0, T] \to \mathbb{R}$ is the scalar-valued diffusion term, and $w_t \in \mathbb{R}^d$ denotes a standard Wiener process.

A common instantiation of the diffusion process is the \emph{Ornstein–Uhlenbeck (OU) process}~\citep{oksendalStochasticDifferentialEquations1995}, in which the drift pulls $x_t$ toward a mean $\mu \in \mathbb{R}^d$:
\begin{equation}
    \mathrm{d}x_t = \beta(t)(\mu - x_t)\,\mathrm{d}t + g(t)\,\mathrm{d}w_t, \quad t \in [0, T],
    \label{eq:ou_process}
\end{equation}
where $\beta \left(t\right)$ is a positive scalar-valued function controlling the drift strength. In this case, the OU process admits a closed-form solution for the mean and variance of the marginal distribution of $x_t$. Specifically, letting
\begin{equation*}
    \bar{\beta}_t := \exp\left(-\int_0^t \beta \left(\nu\right)~\mathrm{d}\nu\right),
\end{equation*}
then the marginal distribution of $x_t$ is Gaussian:
\begin{equation}
    p_t \left(x_t \mid x_0\right) = \mathcal{N}\left(x_t;~\mu - (\mu - x_0)\bar{\beta}_t,~\frac{g\left(t\right)^2}{2\beta(t)}\left(\mathbb{I} - \bar{\beta}_t^2 \mathbb{I}\right)\right),
    \label{eq:distribution}
\end{equation}
where $\mathbb{I} \in \mathbb{R}^{d \times d}$ is the identity matrix.

\paragraph{Sampling Process.} To generate new data, one samples from the reverse-time SDE corresponding to the forward process. Under suitable regularity conditions, this reverse SDE takes the form~\citep{andersonReversetimeDiffusionEquation1982}:
\begin{equation}
    \mathrm{d} x_t = \left[f\left(x_t, t\right) - g\left(t\right)^2 \nabla_x \log p_t \left(x_t  \right)\right]~\mathrm{d}t + g\left(t\right)~\mathrm{d}\widetilde{w}_t, \quad t \in \left[0, T\right], \quad x_T\sim p_{x_T},
    \label{eq:ou_backward}
\end{equation}
where $\widetilde{w}_t$ is a standard Wiener process running backward in time, and $\nabla_x \log p_t(x_t)$ is the score function of the marginal distribution.

In practice, the score function is unknown and approximated by a neural network $s_\theta(x_t, t)$ trained using denoising score matching. The training objective minimizes the expected squared error between the predicted score and the true score:
\begin{equation*}
    \mathbb{E}_{t \sim \mathrm{U} \left[0, T\right]} \mathbb{E}_{x_0 \sim p_0(x)} \mathbb{E}_{x_t \sim p_t \left(x_t \mid x_0\right)} \left[\left| s_\theta \left(x_t, t\right) - \nabla_x \log p_t \left(x_t \mid x_0\right) \right|^2\right],
\end{equation*}
where $\left|\cdot\right|$ denotes the $\ell_2$-norm, and $\mathrm{U} \left[0, T\right]$ the uniform distribution with support $[0,T]$.

\subsection{Constraint-Aware Diffusion Guidance}
\label{sec:constraint_aware}
Before introducing the proposed CoDiG framework, we specify the functional forms of the drift and diffusion terms in~\eqref{eq:ou_process} for concreteness and clarity. It is important to emphasize that the proposed framework does not rely on these specific choices - the following definitions are adopted purely for illustrative purposes and to remain consistent with the experimental setup described later.

We let $\mu = 0$, and define the drift term and the diffusion term as 
\begin{equation*}
    f\left(x_t, t\right) = - \beta \left(t\right) x_t, \quad g\left(t\right) = \sqrt{2 \beta \left(t\right)}, \quad t \in \left[0, T\right],
\end{equation*}
which yields the so-called variance preserving SDE~\citep{songScoreBasedGenerativeModeling2021}, where $g\left(t\right)^2 = 2 \beta \left(t\right)$ holds for all $t\in\left[0, T\right]$ such that the marginal variance of $x_t$ is preserved over time. This specific choice ensures that the forward process remains stable and tractable for training and sampling, while still allowing for an expressive and well-defined reverse-time generative process. Under this formulation, when the terminal time $T$ is sufficiently large, the forward diffusion process described by~\eqref{eq:ou_process} converges to a standard Gaussian distribution. As analyzed in \citet{songScoreBasedGenerativeModeling2021}, the term $\sqrt{2 \beta(t)}$ should grow with time, requiring $\beta(t)$ to be strictly increasing.

For simplicity and numerical stability, we normalize the diffusion process to $t \in \left[0, 1\right]$. To ensure convergence to a standard Gaussian, the diffusion term $\sqrt{2\beta\left(t\right)}$ must grow rapidly within this interval. In our implementation, we model $\beta\left(t\right)$ as a quadratic function, $\beta\left(t\right) = r_1 t^2 + r_0$, with parameters detailed in Appendix~\ref{sec:training_network}. In this case, \eqref{eq:ou_backward} can be reformulated as:
\begin{equation}
    \mathrm{d} x_t = \left[-\beta \left(t\right) x_t -2 \beta \left(t\right) \nabla_x \log p_t \left(x_t \right)\right]~\mathrm{d} t + \sqrt{2 \beta \left(t\right)}~\mathrm{d}\widetilde{w}_t, \quad t \in \left[0, 1\right], \quad x_1\sim p_{x_1}.
    \label{eq:new_backward}
\end{equation}

Next, we consider the marginal distribution $p_t\left(x_t \right)$, which represents the probability distribution of a sample at an intermediate time step, in the absence of constraints. Before incorporating constraints into this distribution, we first introduce the following definitions. Let $c : \mathbb{R}^d \times \left[0,1\right] \rightarrow \mathbb{R}^k$ denote a time-dependent constraint function, encoding the safety or feasibility requirements of the system. We define the feasible region at time $\tau\geq 0$ as
\begin{equation*}
\mathcal{C}_\tau := \left\{ x \in \mathbb{R}^d \mid c\left(x, \tau\right) \leq 0 \right\},   
\end{equation*} 
where the inequality is interpreted element-wise. Naturally, when constraints are introduced, the distribution of interest becomes the conditional distribution:
\begin{equation*}
    p_t \left(x_t \mid \mathcal{C}_{\tau}\right), \quad t\in\left[0, 1\right], \quad \tau \geq 0.
\end{equation*}
These constraints may encode different forms of feasibility or safety requirements, depending on the task setting. For example, in autonomous racing, $\mathcal{C}_\tau$ refers to the obstacle-free, drivable region of a racing track. While in diffusion-based control policies, $\mathcal{C}_\tau$ must account for system dynamics, as the use of visual feedback necessitates control that respects the underlying physical constraints of the system. Here, we use the time subscript $\tau$ to emphasize that such constraints can be time-varying, which is often the case in dynamic or interactive environments. For simplicity, and without loss of clarity, we will omit this subscript in the following when no confusion arises.

Several existing methods attempt to directly model the marginal distribution $p_t \left(x_t \mid \mathcal{C} \right)$ by injecting the constraint representation into the diffusion model architecture~\citep{hoClassifierFreeDiffusionGuidance2022}. While effective in big-data domains such as image synthesis, these approaches face significant limitations in the context of robotics: \begin{inparaenum}[(i)]
    \item Learning $p_t \left(x_t \mid \mathcal{C}\right)$ from scratch requires many expert demonstrations that satisfy $\mathcal{C}$, which are often expensive or impractical to collect in robotics.
    \item Since $\mathcal{C}$ is often time-varying and task-specific, models trained on a fixed distribution may fail to generalize to unseen or dynamic constraints at test time.
\end{inparaenum}

To overcome these limitations, we leverage the known structure of the constraint $\mathcal{C}$ during sampling to dynamically guide the generation process. We propose an alternative formulation of the constrained distribution $p_t \left(x_t \mid \mathcal{C}\right)$, which does not require learning the conditional model directly from data:
\begin{equation*}
    p_t \left(x_t \mid \mathcal{C}\right) = p_t \left(x_t \right) \frac{e^{- \gamma_t V \left(x_t;~\mathcal{C}\right)} }{Z_t},
\end{equation*}
where $Z_t := \int_{\mathbb{R}^d} p_t \left(x \right) e^{- \gamma_t V \left(x,~\mathcal{C}\right)}~\mathrm{d} x$ is a normalization constant. The barrier function $V: \mathbb{R}^d \rightarrow \mathbb{R}_{+}$ assigns large values to infeasible data points, while remaining close to zero within the feasible region. Intuitively, applying the barrier function pushes the probability of infeasible data points toward zero. As a result, the constrained distribution focuses its support almost entirely on the feasible region. We substitute the above formula into the score function and get
\begin{equation}
    \nabla_x \log p_t \left(x_t \mid \mathcal{C}\right) = \nabla_x \log p_t \left(x_t \right) - \gamma_t \nabla_x V \left(x_t;~\mathcal{C}\right),
    \label{eq:new_score_function}
\end{equation}
where the normalization constant $Z_t$ vanishes when taking the gradient of the log-probability, and hence does not affect the reverse-time dynamics. By substituting the right-hand side of~\eqref{eq:new_score_function} into~\eqref{eq:new_backward}, we obtain the modified reverse SDE that incorporates constraint information:
\begin{equation}
    \mathrm{d} x_t = \beta \left(t\right) \left[- x_t - \left(1+\eta\right) \left(\nabla_x \log p_t \left(x_t\right)  - \gamma_t \nabla_x V \left(x_t;~\mathcal{C} \right) \right)\right]~\mathrm{d} t + \eta \sqrt{2 \beta \left(t\right)}~\mathrm{d}\widetilde{w}_t,
    \label{eq:constraint_backward}
\end{equation}
where a constant $\eta \in \left[0, 1\right]$ is introduced to accelerate convergence and enhance the stability of the sampling process~\citep{songImprovedTechniquesTraining2020}. We observe that the first term on the right-hand side of~\eqref{eq:new_score_function} corresponds exactly to the unconstrained score function defined in~\eqref{eq:new_backward}. This term can still be approximated by the neural network $s_{\theta} \left(x_t, t\right)$ trained without considering any constraints. Crucially, the effect of the constraint appears only during the denoising process, in an explicit gradient-based form - as an additive term derived from the constraint potential (e.g., a barrier function). This formulation significantly reduces the need for large amounts of constraint-compliant training data, as the constraint is not encoded in the network itself but \emph{instead injected at inference time}. Moreover, because the constraint enters the reverse SDE as a differentiable time-varying potential, the framework can naturally accommodate dynamic, time-varying constraints.

It is important to note, as pointed out by \citet{bastekPhysicsInformedDiffusionModels2025}, that applying constraints to data that is close to pure noise in diffusion models is not meaningful. Therefore, during the denoising process - i.e., as $t$ decreases from one to zero - we gradually increase the value of $\gamma_t$ starting from zero at $t=1$. This progressive scheduling is crucial for ensuring the stability of the denoising process. The specific design of $\gamma_t$ is detailed in Appendix~\ref{sec:hyperparameter}.

\section{Case Study of Autonomous Racing}
\label{sec:autonomous_racing}
In this section, we illustrate how to design a constraint-aware barrier function and analyze its impact on inference through a concrete application - obstacle avoidance in autonomous racing. While this example serves to ground our discussion, the barrier function design and constraint-handling mechanisms are task-agnostic. Thus, our framework is not limited to autonomous racing but serves as a general-purpose solution for safety-critical robotics applications. For details on dataset construction, diffusion model architecture, and training procedures, please refer to Appendix~\ref{sec:data_arch_train}.

\subsection{Constraint-Aware Barrier Function}
\label{sec:barrier_function}
For the considered application, the barrier function is designed as follows:
\begin{equation}
    V\left(\hat{y},\widehat{\phi};~\mathcal{C}\right) = \sum_{k=0}^{N-1}~ \underbrace{\alpha \mathds{1}\left\{\hat{y}_k \notin \mathcal{C}_{k} \right\}}_{\text{first part}} + \underbrace{\frac{\epsilon}{2} \left|\hat{y}_k - \hat{y}_{\text{nominal},k}\right|^2 + \frac{\epsilon}{2} \left|\widehat{\phi}_k - \widehat{\phi}_{\text{nominal}, k}\right|^2}_{\text{second part}},
    \label{eq:barrier_function}
\end{equation}
where the symbol $\mathds{1}\left\{\cdot\right\}$ denotes the indicator function, and the subscript $\left(\cdot\right)_{\text{nominal}}$ refers to the time-optimal solution computed offline in the absence of obstacles, which serves as a reliable reference. Here, $N$ represents the number of discrete points obtained by uniformly sampling along the track center line. In our setting, $\hat{y}$ denotes the lateral displacement and $\widehat{\phi}$ represents the yaw angle in the Frenet coordinate system (see Appendix~\ref{sec:data_construction}). The feasible region $\mathcal{C}_k$ is also defined in the Frenet frame, capturing the obstacle-free area at each sampled position along the track.

In~\eqref{eq:barrier_function}, the first part accounts for time-varying obstacles by guiding the sampling process toward safe, obstacle-free regions. The second part addresses the missing curvature information of the track during training, which is intentionally discarded when transforming trajectories into the local Frenet coordinate system. Beyond promoting near time-optimality without requiring global geometric knowledge of the track, this part also facilitates the generation of dynamically feasible trajectories. The local representation ensures that the resulting motions adhere more closely to the physical and kinematic constraints of the system. The positive constants $\alpha$ and $\epsilon$ are tunable hyperparameters that balance the importance of the two components. Specifically, $\alpha$ modulates the influence of physical safety constraints, while $\epsilon$ regulates the adherence to nominal time-optimality.

It is worth noting that the design of the barrier function is not unique and can be tailored to the specific task. While such customization may require a small amount of tuning effort, it is negligible compared to the cost of collecting expert demonstrations, especially in robotics domains where data is expensive. This makes our framework both flexible and data-efficient.

\subsection{Constraint-Aware Inference}
\label{sec:inference}
\begin{figure}[htbp]
  \centering
  \begin{subfigure}[t]{\textwidth}
    \centering
    \includegraphics[width=\linewidth]{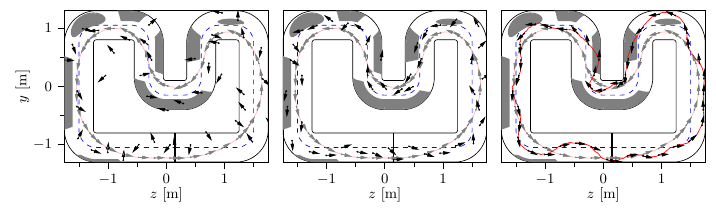}
    \caption{Sampling without barrier function.}
    \label{fig:without_barrier}
  \end{subfigure}
  \vspace{1em}
  \begin{subfigure}[t]{\textwidth}
    \centering
    \includegraphics[width=\linewidth]{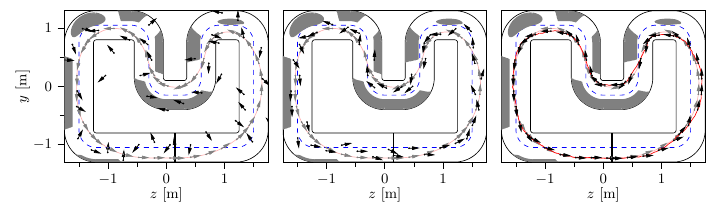}
    \caption{Sampling with barrier function.}
    \label{fig:with_barrier}
  \end{subfigure}
  \caption{Intermediate denoising results during sampling at three representative time steps \(t = \qty{1}{\second},~\qty{0.591}{\second},~\qty{0.002}{\second}\), from left to right. (a) Sampling process without the barrier function. (b) Sampling process with the proposed barrier function. Black dots and arrows represent the generated trajectory points and their heading directions in the global frame.}
  \label{fig:barrier_comparison}
\end{figure}

We train the diffusion model as described in Appendix~\ref{sec:network} for $\num{500}$ epochs. During inference, we applied the Euler-Maruyama discretization~\eqref{eq:euler_maruyama}, which corresponds to the discretized version of~\eqref{eq:constraint_backward}. The denoising process proceeds from $t=1.0$ to $t = 0.0$ in $\num{1000}$ steps, gradually transforming samples from noise to data. The results are shown in Fig.~\ref{fig:barrier_comparison}. Fig.~\ref{fig:without_barrier} illustrates the denoising process without using the barrier function, while Fig.~\ref{fig:with_barrier} shows the effect of the proposed constraint-aware guided generation. Each row depicts intermediate generation results at \(t = \qty{1}{\second},~\qty{0.591}{\second},~\qty{0.002}{\second}\), from left to right. The black points and arrows indicate the evolving trajectories in the $z$-$y$ plane. See Appendix~\ref{sec:hyperparameter} for the concrete values of the hyperparameters used during inference.

In Fig.~\ref{fig:without_barrier}, we observe that the diffusion model learns important features of the data distribution. It respects the fundamental constraints of the racetrack, such as staying within bounds and satisfying the loop closure. However, despite conditioning on constraints during training, the generated samples still fail to consistently avoid obstacles - primarily due to limited training data. Additionally, the lack of curvature information in the local Frenet frame leads to unrealistic and physically implausible results.

In contrast, Fig.~\ref{fig:with_barrier} incorporates the barrier function as described in Sec.~\ref{sec:barrier_function}. The guidance significantly improves the sampling process. The model denoises faster (i.e., the samples become structured earlier), the trajectories successfully avoid all obstacles, and the resulting motion aligns well with a physically consistent motion. Moreover, due to the second part in~\eqref{eq:barrier_function} (albeit with a small weight $\epsilon$), the final trajectories closely follow the nominal time-optimal ones, exhibiting near time-optimal properties. For more details on the near time-optimality of the generated trajectories, please refer to Appendix~\ref{sec:near_optimal}.

\section{Real-World Experiments}
\label{sec:experiments}
We evaluate CoDiG in experiments on a real-world miniature autonomous racing platform~\citep{bodmerOptimizationBasedSystemIdentification2024,carronChronosCRSDesign2023}. For more details on the experimental platform, the obstacle setup, and a flowchart illustrating how the CoDiG framework is deployed to achieve real-time obstacle avoidance, please refer to Appendix~\ref{sec:structure}.

\subsection{Warm-Starting}
Real-time obstacle avoidance requires not only safe trajectories but also fast replanning. As shown in Sec.~\ref{sec:inference}, our diffusion model with a barrier function produces high-quality trajectories after $\num{1000}$ denoising steps, but this results in a low sampling frequency of $\qty{0.25}{\hertz}$, which is insufficient for real-time racing.

While various acceleration techniques exist~\citep{songDenoisingDiffusionImplicit2022,paloDiffusionAugmentedAgents2025,zhangFastSamplingDiffusion2023}, we propose a warm-start strategy tailored to robotic control. Unlike standard diffusion generation, which samples each trajectory from pure noise, our proposed warm-start technology perturbs the previous output with small noise and reuses it as the next input. This maintains temporal consistency, reduces trajectory variance, and improves control stability~\citep{ModelPredictiveControl1999}. By promoting smooth transitions between consecutive trajectories, warm-starting significantly lowers the number of denoising steps required and enhances real-time feasibility. A detailed analysis and comparison of results with and without warm-starting are provided in Appendix~\ref{sec:warm_start}.

\subsection{Experimental Results}
Through the integration of our warm-start technique, we achieve a sampling frequency of $\qty{2.5}{\hertz}$ on a computer equipped with an \verb|NVIDIA RTX 4090| GPU. While there is still potential for further acceleration, this performance already satisfies the real-time requirements of obstacle avoidance in racing scenarios.

\begin{figure}[htbp]
  \centering
  \begin{subfigure}[t]{\textwidth}
    \centering
    \includegraphics[width=\linewidth]{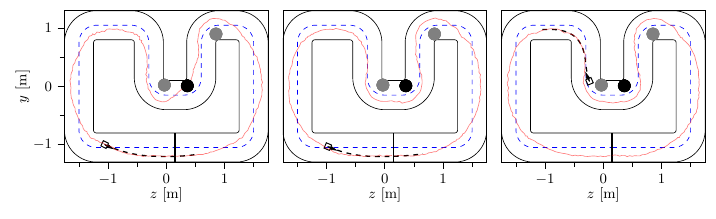}
    \caption{Obstacle avoidance episode 1.}
    \label{fig:real_demo1}
  \end{subfigure}
  \vspace{1em}
  \begin{subfigure}[t]{\textwidth}
    \centering
    \includegraphics[width=\linewidth]{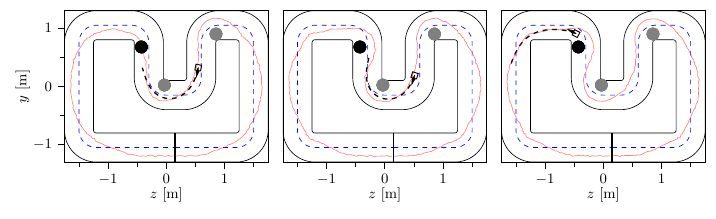}
    \caption{Obstacle avoidance episode 2.}
    \label{fig:real_demo2}
  \end{subfigure}
  \caption{Real-world demonstration of real-time obstacle avoidance using CoDiG. Red lines represent the planned trajectories generated by the CoDiG diffusion planner. Gray circles indicate static obstacles, and black circles represent dynamic obstacles. Black dashed lines show the predicted trajectory from the TMPC while following the reference plan. Each episode illustrates a complete avoidance cycle: obstacle encroachment, real-time replanning, and successful passage.}
  \label{fig:demo}
\end{figure}

We successfully deployed CoDiG on our real-world autonomous racing platform for real-time trajectory planning. A tracking model predictive control (TMPC)~\citep{MPCTrackingPiecewise2008,solopertoNonlinearMPCScheme2023} is employed to follow the planned trajectories. Notably, the TMPC operates without any knowledge of obstacles, relying solely on the reference trajectories for control. Fig.~\ref{fig:demo} illustrates two representative obstacle avoidance maneuvers during autonomous driving. In both Fig.~\ref{fig:real_demo1} and Fig.~\ref{fig:real_demo2}, the red lines represent the trajectories planned by CoDiG, the gray circles denote static obstacles, while the black circles indicate dynamic obstacles. The black dashed lines show the predicted trajectories generated by the TMPC as it attempts to follow the red reference trajectory. Each sequence from left to right captures a complete avoidance cycle: 
\begin{inparaenum}[(i)]
    \item Obstacle Encroachment: An obstacle intrudes into a previously feasible trajectory, making it infeasible.
    \item Replanning: The planner detects the encroachment and generates a new collision-free trajectory.
    \item Successful Avoidance: The vehicle safely bypasses the obstacle.
\end{inparaenum}

As shown in the figures, the predicted trajectories from the TMPC closely align with the reference trajectories generated by the diffusion model. This highlights that the planned trajectories are closely aligned with physical feasibility, enabled by the barrier function, which is crucial for effective tracking performance. Additionally, even in the presence of obstacles, the generated trajectories maintain near time-optimality, indicating that the planner does not overly sacrifice efficiency for safety.

Finally, thanks to the warm-start strategy, significant replanning is only triggered when the obstacle actually interferes with the current path. In static conditions, consecutive trajectories remain almost unchanged, ensuring system stability. We conducted five experimental trials, each consisting of $\num{15}$ racing laps, across ten different obstacle configurations. The framework achieved a $\qty{100}{\percent}$ success rate in obstacle avoidance, demonstrating its robustness and reliability in diverse scenarios.

\section{Conclusion}
\label{sec:conclusion}
In this work, we propose CoDiG (Constraint-Aware Diffusion Guidance), a general, data-efficient framework that leverages diffusion models for real-time, safety-critical motion planning in robotics. While diffusion models have shown strong capabilities in learning complex distributions, their direct application in robotics is hindered by the lack of constraint-awareness and physical feasibility. We address this challenge by introducing a barrier function into the denoising process, guiding the generated trajectories toward safe and dynamically consistent regions without requiring extensive expert data. A warm-start inference strategy further improves inference speed and temporal consistency for real-time deployment.

We demonstrate CoDiG on autonomous racing with dynamic obstacles, achieving robust real-world performance with reliable obstacle avoidance, precise tracking, and near time-optimal planning at $\qty{2.5}{\hertz}$. These results highlight the potential of diffusion-based methods for constraint-aware planning and control, offering a promising direction for safe, efficient, and generalizable robotic decision-making in varying environments.

\acknowledgments{We thank Matteo Facchino for providing code related to time-optimal control solvers. We also gratefully acknowledge Jan-Hendrik Bastek for the insightful discussions on constraint handling in diffusion models. We thank the German Research Foundation and the Max-Planck ETH Center for Learning Systems for the support.}


\bibliography{example}  

\clearpage

\appendix

\section{Hyperparameters}
\label{sec:hyperparameter}
During inference, the hyperparameters are set as follows:
\begin{equation*}
    \eta = 0.1, \quad \alpha = 0.4, \quad \epsilon = 16.0.
\end{equation*}
In particular, the time-varying weight $\gamma\left(t\right)$ is assigned non-uniform values according to the following scheme:
\begin{equation*}
    \gamma\left(t\right) = \frac{\hbar_1}{1 + \exp{\left(-\hbar_2 \left(\hbar_3 - t\right)\right)}},\quad t\in \left[0, 1\right],
\end{equation*}
where $\hbar_1=1.0$, $\hbar_2=50.0$, and $\hbar_3=0.7$.

\section{Discrete-Time Integration}
\label{sec:discrete_sde}
Assuming a denoising process over $M \in \mathbb{N}_{++}$ steps, we partition the interval $\left[0, 1\right]$ non-uniformly as follows:
\begin{equation*}
    t_k = \left(1 - \frac{k}{M}\right)^p,\quad k=0,\dots,M,
\end{equation*}
where $p=2.2$ in our case. Starting from an initial sample $x_0$ drawn from a standard Gaussian distribution, we perform denoising according to the following discrete Euler-Maruyama~\citep{sauerNumericalSolutionStochastic2012} update scheme:
\begin{equation}
\begin{aligned}
    \bar{x}_{k+1} &= x_k +
    \beta \left(t_k\right) \left[- x_k - \left(1+\eta\right) \left( s_{\theta}\left(x_k, t_k\right)  - \gamma_{t_k} \nabla_x V \left(x_k;~\mathcal{C}\right) \right)\right]~\Delta t_k,  \\
    x_{k+1} &= \bar{x}_{k+1} + \eta \sqrt{2 \beta \left(t_k\right)}~\sqrt{\left|\Delta t_k\right|} \sigma_k,\quad k=0,\dots, M-1,
\end{aligned}
\label{eq:euler_maruyama}
\end{equation}
where $\Delta t_k = t_{k+1} - t_k$ denoting the step size between successive time points. The noise term $\sigma_k \in \mathbb{R}^{d}$ is sampled from a standard Gaussian distribution. Here, $\bar{x}$ denotes the mean estimate at each step, while $x$ denotes the noisy sample.

\section{Data and Model Pipeline}
\label{sec:data_arch_train}
In this section, we describe the pipeline used to train our diffusion model for obstacle avoidance in racing scenarios. We begin by presenting our data collection process, where expert demonstrations are gathered to reflect optimal driving behaviors in the presence of obstacles. Then, we introduce a data augmentation strategy that diversifies the training distribution while preserving expert intent. Next, we detail the architecture of our proposed diffusion model, which is adapted to handle time-conditioned inputs and spatial constraints relevant to the racing task. Finally, we present the training results of the diffusion model under various input configurations, demonstrating how different modalities affect the training performance.

\subsection{Learning-Efficient Dataset Construction}
\label{sec:data_construction}
Even on a miniature autonomous racing platform, collecting expert demonstrations via manual teleoperation is highly challenging and time-consuming. Therefore, we generate expert data by solving a time-optimal control problem~\citep{verschueren2016time}, including car states and control inputs.

\begin{figure}[htbp]
  \centering
  \begin{subfigure}[t]{0.48\textwidth}
    \centering
    \includegraphics[width=\linewidth]{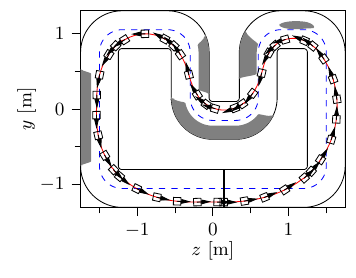}
    \caption{Time-optimal expert trajectory.}
    \label{fig:track_obs1}
  \end{subfigure}
  \hfill
  \begin{subfigure}[t]{0.48\textwidth}
    \centering
    \includegraphics[width=\linewidth]{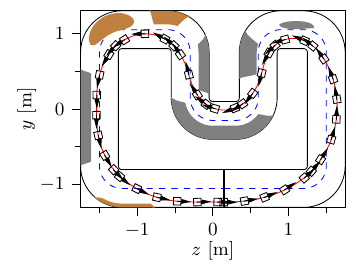}
    \caption{Redundant obstacle augmentation.}
    \label{fig:track_obs2}
  \end{subfigure}
  \vspace{1em}
  \begin{subfigure}[t]{0.98\textwidth}
    \centering
    \includegraphics[width=\linewidth]{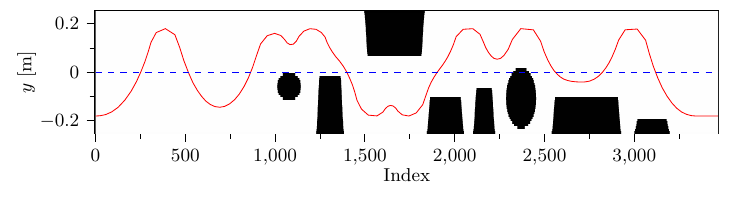}
    \caption{Flattened Frenet representation.}
    \label{fig:flatten_map}
  \end{subfigure}
  \caption{(a) A time-optimal trajectory (red line) computed for a given obstacle configuration (gray regions) on the racing track. Black rectangles and arrows indicate the approximate vehicle shape and heading. (b) Redundant obstacles (brown regions) added in areas that do not affect the trajectory, providing data augmentation without solving additional optimal control problems. (c) A flattened track representation in the local Frenet coordinate system, visualizing both the trajectory (red line) and obstacles (black regions).}
  \label{fig:track_and_obs}
\end{figure}

To collect expert data, we randomly place obstacles on the track and solve the aforementioned time-optimal control problem to obtain optimal driving trajectories with continuous looping and corresponding control inputs. An example is shown in Fig.~\ref{fig:track_obs1}, where the gray regions indicate obstacles. The red curve shows the trajectory in the $z$-$y$ plane, and the black rectangles and arrows illustrate the approximate shape and orientation of the vehicle, respectively, reflecting the fact that the vehicle is not treated as a point mass to account for the system dynamics.

As previously mentioned, collecting expert data is expensive. Solving a single time-optimal control problem takes around $\num{10}$ minutes on average. To address this limitation, we propose a method for dataset augmentation. We observe that once the time-optimal solution is obtained for a given map (with a specific obstacle configuration), adding extra obstacles within the safe region that do not interfere with the trajectory will not alter the time-optimal solution. These redundant obstacles - illustrated as brown regions in Fig.~\ref{fig:track_obs2} - can be arbitrarily placed without affecting the outcome. Based on this observation, we first collect $\num{100}$ trajectories by solving time-optimal problems with randomly placed obstacles, which takes approximately $\num{16}$ hours in total. We then expand this dataset to $\num{10000}$ trajectories by adding random redundant obstacles in safe regions, using $\qty{80}{\percent}$ of them when training the diffusion model.

During training, we only use the pose information - namely $y$ and yaw angle $\phi$ - which are transformed into a local Frenet coordinate system~\citep{crenshawOrientationHelicalMotion1993}. This yields the local variables $\hat{y}$ and $\hat{\phi}$, representing the lateral displacements and heading relative to the reference path. Together with the obstacle representation, this results in a flattened map as shown in Fig.~\ref{fig:flatten_map}. In this map, the presence of obstacles naturally induces an obstacle-free region, denoted by $\mathcal{C}$, which is already defined in the local Frenet frame. For notational simplicity, we omit the explicit time index $\tau$, but we emphasize that $\mathcal{C}$ is inherently time-varying, reflecting the dynamic nature of the environment. The set $\mathcal{C}$ provides a time-varying constraint in the planning process and is considered in the definition of our constraint-aware barrier function.

By performing this transformation, we deliberately discard information about the global curvature of the track. This enhances the generalization capability of the trained diffusion model, enabling the model to handle arbitrary (even moving) obstacles. However, this also means that the generated trajectories may not inherently account for curvature constraints, an issue we address using a barrier function in the denoising process, which is detailed in Sec.~\ref{sec:inference}. 

\subsection{Diffusion Model Architecture}
\label{sec:network}
\begin{figure}[htbp]
  \centering
  \includegraphics[width=\textwidth]{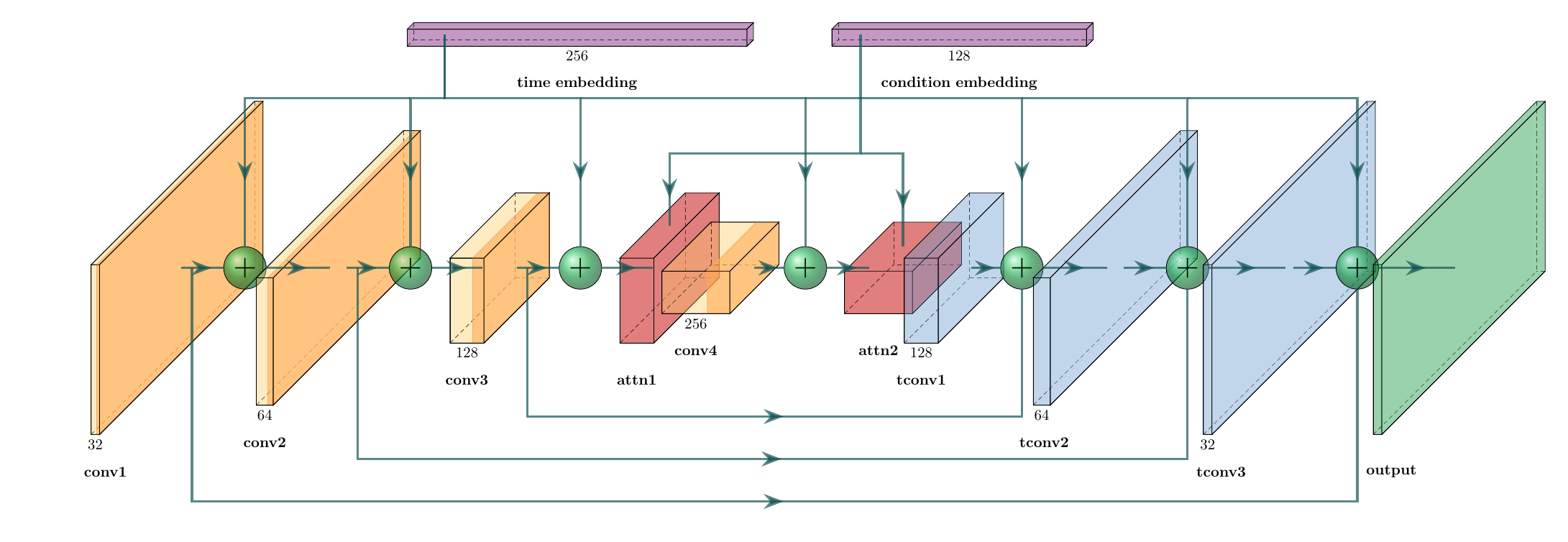} 
  \caption{Architecture of the proposed time-conditioned score-based generative model. The U-Net backbone extracts multi-scale features through a sequence of convolutional and deconvolutional layers, with temporal embeddings injected via dense layers. Spatial transformer modules enable conditional attention guided by task-specific context. Skip connections ensure spatial consistency across scales.}
  \label{fig:unet}
\end{figure}

As illustrated in Fig.~\ref{fig:unet}, we adopt a time-conditioned U-Net architecture as the backbone of our diffusion model~\citep{rombachHighResolutionImageSynthesis2022}. The network follows a classic encoder-decoder structure, augmented with time and conditional information to support trajectory generation in dynamic environments.

The input is a single-channel spatial-temporal representation of the trajectory, and the output preserves the same spatial resolution. Temporal conditioning is achieved via Gaussian Fourier features~\citep{hensmanVariationalFourierFeatures}, which embed the diffusion time step into a high-dimensional representation. This embedding is injected at every resolution level to inform the network of the denoising progress.

The encoder consists of a sequence of down-sampling convolutional blocks, each followed by time embedding fusion and group normalization. To enhance spatial reasoning and enable conditional generation, spatial transformer modules are inserted at deeper layers, where they incorporate context information - such as a reference track - encoded via a lightweight convolutional neural network.

The decoder mirrors the encoder with up-sampling blocks and skip connections, allowing the network to reconstruct high-resolution outputs by fusing low-level and high-level features. Each decoding layer is also conditioned on time to ensure consistency with the diffusion process.

This architecture is designed to be data-efficient, modular, and generalizable. It supports plug-and-play conditional guidance and is easily extendable to other tasks in robotics beyond the case study of autonomous racing.

\subsection{Training of the Network}
\label{sec:training_network}
\begin{figure}[htbp]
  \centering
  \includegraphics[width=\textwidth]{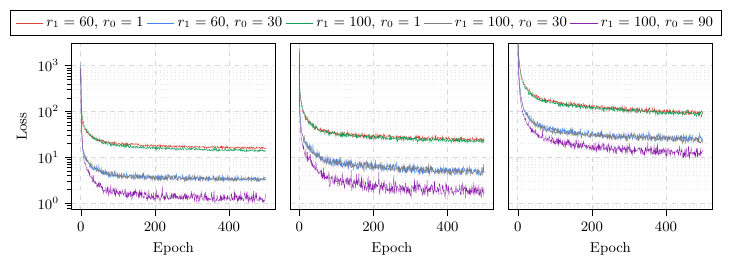} 
  \caption{Training performance of the diffusion model under different input configurations and noise schedules. From left to right, the three plots correspond to using (i) lateral displacement $\hat{y}$ in the Frenet frame only, (ii) lateral displacement $\hat{y}$ and yaw angle $\widehat{\phi}$ in the Frenet frame, and (iii) the states $\hat{x},~\hat{y},~\widehat{\phi}$ along with velocities $\hat{v}_x,~\hat{v}_y,~\widehat{\omega}$ as model inputs. Each setting was trained for 500 epochs while varying the parameters $r_1$ and $r_0$ in the noise schedule.}
  \label{fig:train_loss}
\end{figure}

We experiment with different input modalities for the diffusion model. Specifically, we considered: 
\begin{inparaenum}[(i)]
    \item the lateral displacement $\hat{y}$ after transforming into the Frenet coordinate system;
    \item both the lateral displacement $\hat{y}$ and the yaw angle $\widehat{\phi}$ in the Frenet frame;
    \item the states including $\hat{x},~\hat{y},~\widehat{\phi}$ along with their corresponding velocities $\hat{v}_x,~\hat{v}_y,~\widehat{\omega}$ in the Frenet frame.
\end{inparaenum}
For each input configuration, we train the model for $\num{500}$ epochs and explore different values of $r_1$ and $r_0$ in constructing the noise schedule $\beta\left(t\right) = r_1 t^2 + r_0$ for $t\in\left[0, 1\right]$. The training results are shown in Fig.~\ref{fig:train_loss}, where the three plots from left to right correspond to the aforementioned three input configurations, respectively.

Our experiments show that varying $r_1$ has negligible impact on the final training performance. In contrast, increasing $r_0$ generally improves training outcomes, suggesting that larger initial noise levels may facilitate better learning. However, due to the limited size of our training dataset, excessively large values of $r_0$ can lead to overfitting risks. Additionally, we observe that as the input dimensionality increases, the training performance degrades, likely due to the increased complexity of the data distribution and the limited model capacity under fixed training resources. Based on these observations, we choose to use only the lateral displacement $\hat{y}$ and the yaw angle $\widehat{\phi}$ in the Frenet frame as inputs in our final framework, setting $r_1 = 100.0$ and $r_0=30.0$.

It is important to emphasize that although we adopt a simplified input representation in this work, our approach remains general and can naturally extend to handle higher-dimensional or multimodal inputs. This flexibility paves the way toward directly modeling control inputs using diffusion models in future work.

\section{Warm-Start Evaluation}
\label{sec:warm_start}
In this work, we incorporate a warm-starting strategy to accelerate the sampling process, thereby enabling real-time obstacle avoidance. This section presents a quantitative analysis of the effects introduced by this partial diffusion strategy on trajectory generation performance.

\begin{figure}[htbp]
  \centering
  \includegraphics[width=\textwidth]{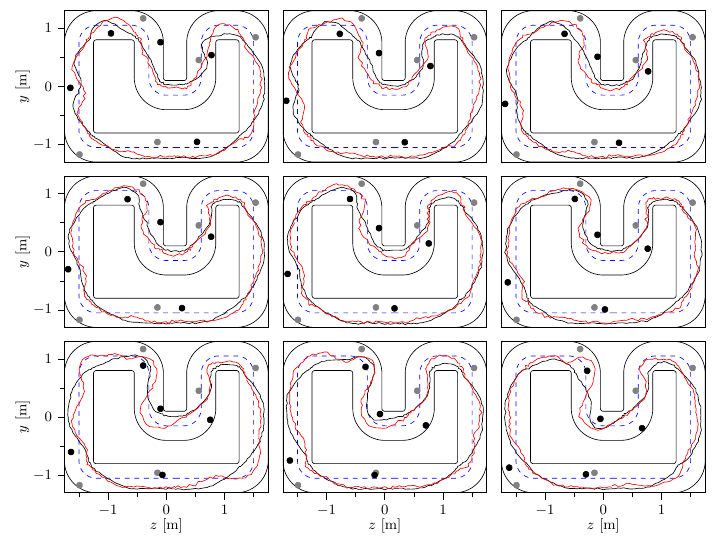} 
  \caption{Comparison of reference trajectories generated with and without the warm start technique under an identical obstacle configuration. Gray circles denote static obstacles, and black circles denote dynamic obstacles. The black solid lines represent trajectories produced by the standard diffusion model after $500$ denoising steps starting from standard Gaussian noise. The red solid lines represent trajectories generated using the warm start approach, where $50$ denoising steps are performed. The warm start method accelerates the sampling process while maintaining successful obstacle avoidance, albeit with slightly coarser trajectory profiles and more conservative motion planning behavior.}
  \label{fig:warm_start}
\end{figure}

Fig.~\ref{fig:warm_start} illustrates the reference trajectories generated with and without the application of the warm start technique under an identical obstacle configuration, sampled at consistent time instances. In the figures, gray circles denote static obstacles, while black circles denote dynamic obstacles. The nine subfigures are arranged sequentially from left to right and top to bottom. In each subfigure, the black solid line represents the trajectory obtained using the standard diffusion model, which initiates from standard Gaussian noise and progresses through $500$ denoising steps. In contrast, the red solid line corresponds to the trajectory generated with the warm start method, which undergoes $50$ denoising steps of partial noised initial trajectory.

As evidenced by the results, both approaches successfully achieve obstacle avoidance at all time steps, demonstrating their respective effectiveness. Nevertheless, the trajectories generated via the warm start technique exhibit a coarser structure, primarily due to the incomplete denoising process inherent to partial diffusion. Furthermore, from the perspective of physical feasibility, the trajectories derived from the standard diffusion model better adhere to realistic vehicle dynamics. Specifically, the final subfigure demonstrates that the warm start method tends to converge to a local solution and favors a more conservative path - remaining closer to the previous time point - to avoid obstacle. Despite this conservatism, the warm start approach proves crucial, as it reduces the sampling time by approximately a factor of three, thereby making real-time obstacle avoidance feasible. Moreover, the conservative behavior introduced by warm start contributes positively to the overall system stability.

\section{Near Time-Optimality}
\begin{figure}[htbp]
  \centering
  \includegraphics[width=\textwidth]{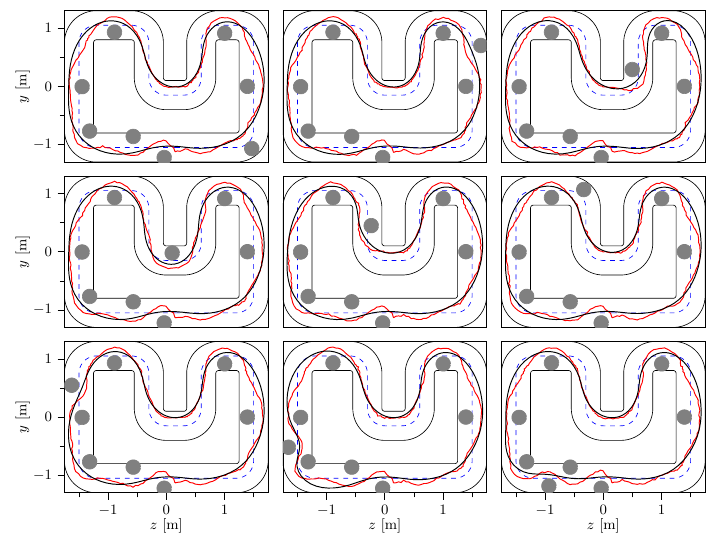} 
  \caption{Comparison between trajectories generated in real time by CoDiG (red) and offline-computed time-optimal trajectories (black) under various obstacle configurations.}
  \label{fig:toc_comparison}
\end{figure}

In this section, we demonstrate the near time-optimality of the trajectories generated by CoDiG by comparing them with trajectories obtained by solving an offline time-optimal control problem~\citep{verschueren2016time}. As illustrated in Fig.~\ref{fig:toc_comparison}, we present several representative obstacle configurations extracted from a real-world experiment. In each scenario, the red trajectory denotes the real-time obstacle-avoidance path generated by the CoDiG framework, while the black trajectory represents the time-optimal path computed offline under the same obstacle layout.
\label{sec:near_optimal}

Overall, we observe a high degree of similarity between the real-time and offline trajectories, which highlights the near time-optimal generation of CoDiG in practice. The main discrepancies are observed in two typical situations. First, to achieve faster cornering, the offline time-optimal solution tends to favor a larger turning radius in curved sections. Second, when navigating near obstacles, the CoDiG-generated trajectory increases its clearance for safety, resulting in a slight deviation from the time-optimal path. This trade-off ensures safety while maintaining strong time-efficiency.

\section{Experimental Platform and the CoDiG Framework}
\label{sec:structure}
\begin{figure}[htbp]
\centering
\begin{subfigure}[t]{0.505\textwidth}
  \centering
  \includegraphics[width=\linewidth]{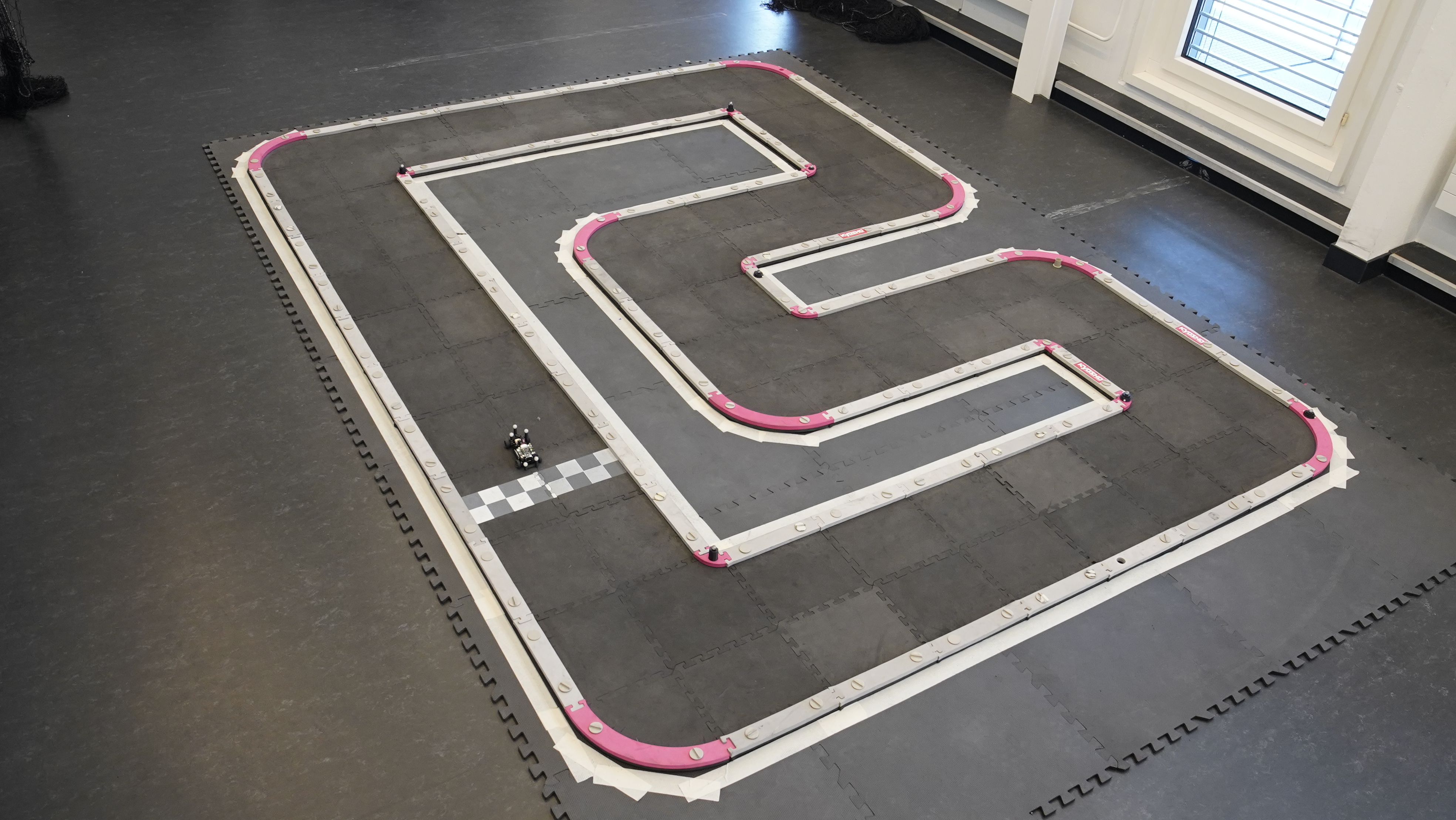}
  \caption{Down-scaled race track.}
  \label{fig:exp_track}
  \end{subfigure}
  \hfill
    \begin{subfigure}[t]{0.455\textwidth}
  \centering
  \includegraphics[width=\linewidth]{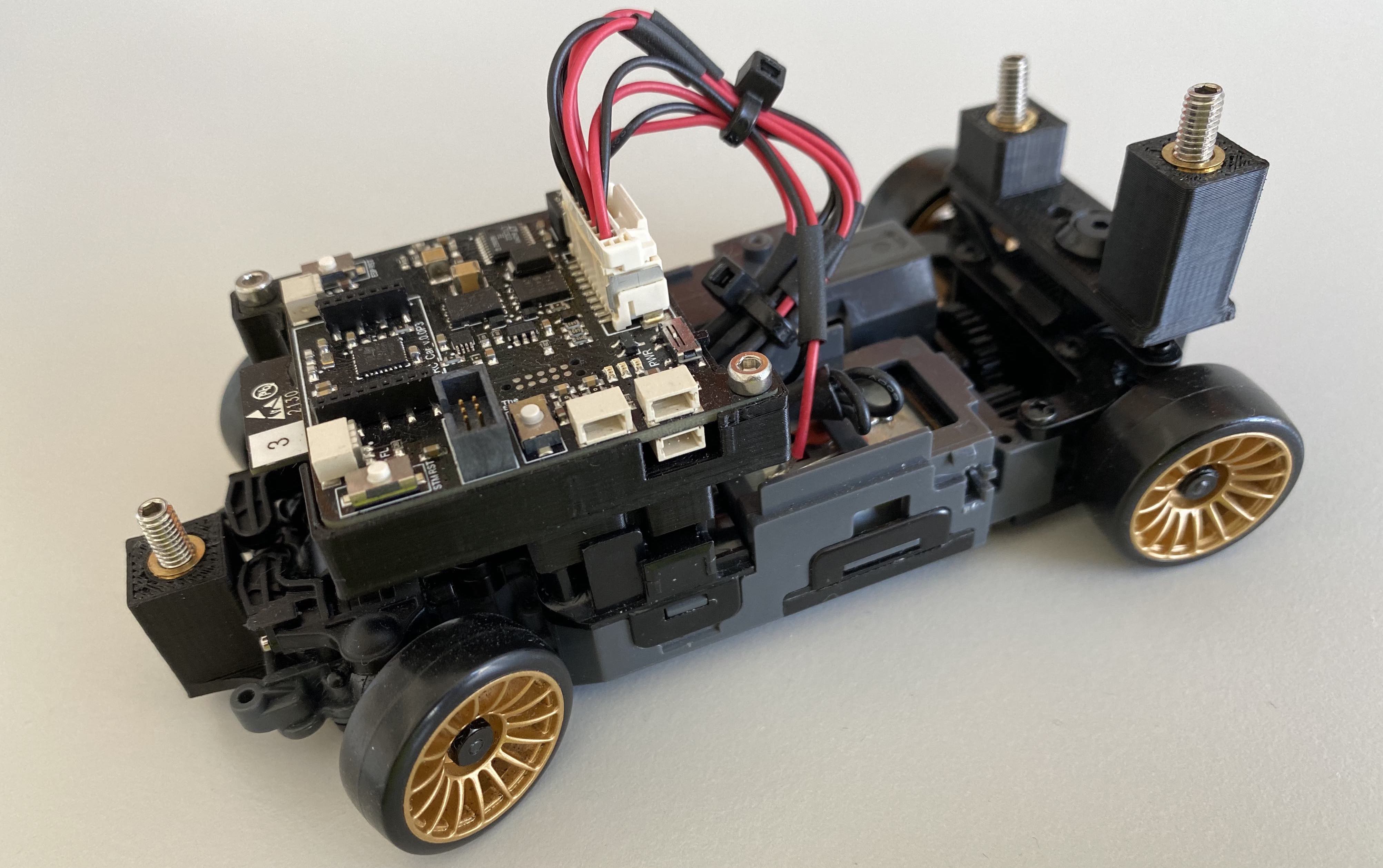}
  \caption{Custom-built autonomous car.}
  \label{fig:exp_car}
  \end{subfigure}
  \vspace{1em}
  \begin{subfigure}[t]{0.98\textwidth}
    \centering
    \includegraphics[width=\linewidth]{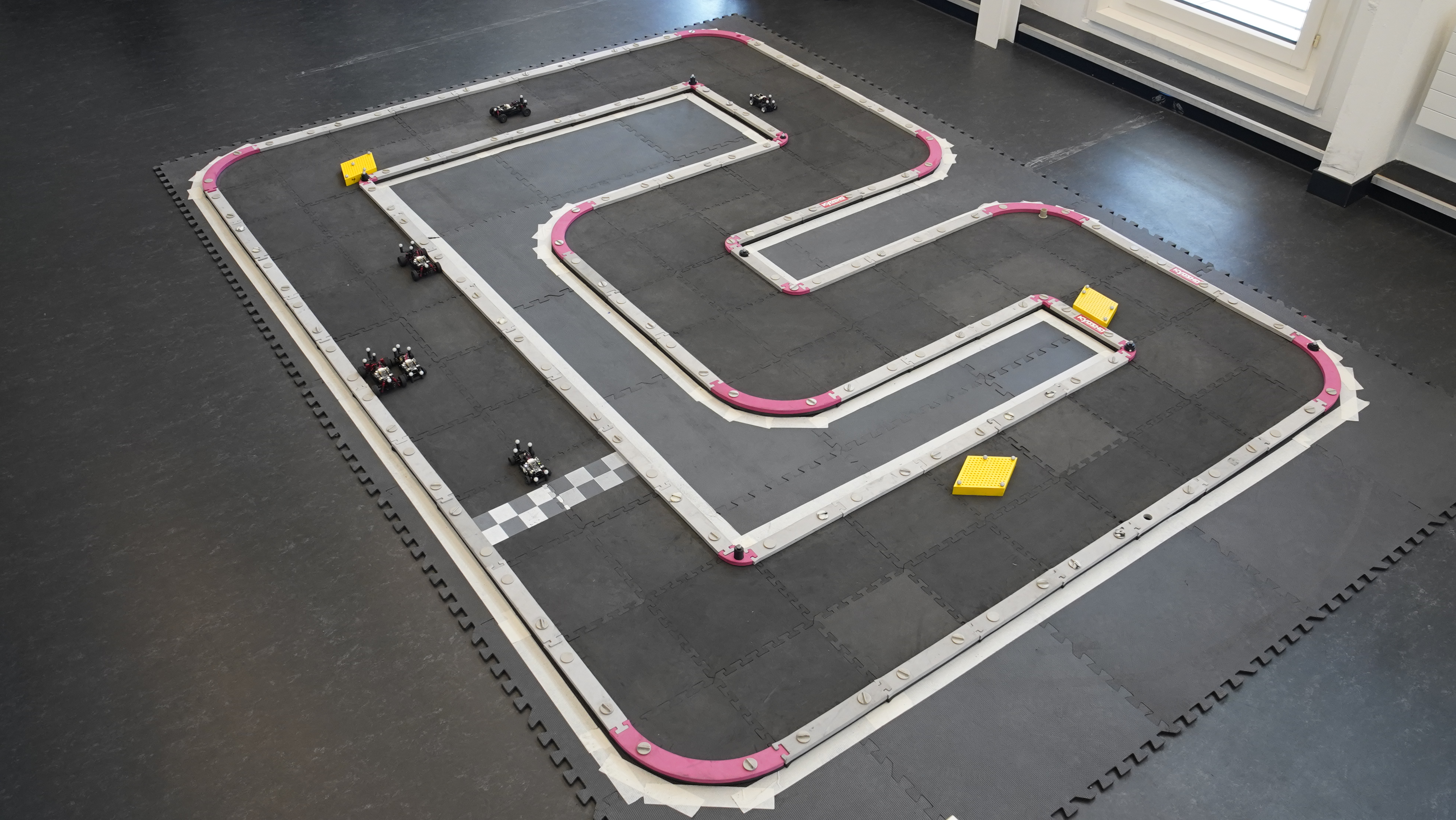}
    \caption{Obstacle configuration.}
    \label{fig:exp_config}
  \end{subfigure}
  \caption{Experimental platform used to evaluate the performance of the CoDiG framework for real-time obstacle avoidance in autonomous racing. The setup includes (a) a down-scaled race track, (b) a custom-built autonomous vehicle, and (c) an obstacle configuration that simulates a challenging and realistic racing scenario.}
  \label{fig:platform}
\end{figure}

Fig.~\ref{fig:platform} illustrates the experimental platform used to evaluate the performance of the CoDiG framework for real-time obstacle avoidance in autonomous racing. The platform consists of a down-scaled race track (Fig.~\ref{fig:exp_track}), a custom-built autonomous car (Fig.\ref{fig:exp_car}), and a motion capture system (not shown in the figure). This setup enables agile maneuvering and real-time control in dynamic, safety-critical scenarios such as obstacle avoidance. It provides a reproducible environment to evaluate our approach under realistic conditions.

Additionally, Fig.~\ref{fig:exp_config} depicts the obstacle configuration used during the experiments. The vehicle positioned at the starting line is the one under our control, responsible for executing the obstacle avoidance task. Yellow boxes represent static obstacles, while the remaining vehicles serve as either dynamic or static obstacles. This setup faithfully simulates a complex and challenging racing environment, emphasizing the effectiveness and robustness of our framework under realistic, high-difficulty conditions.

\begin{figure}[htbp]
  \centering
  \includegraphics[width=\textwidth]{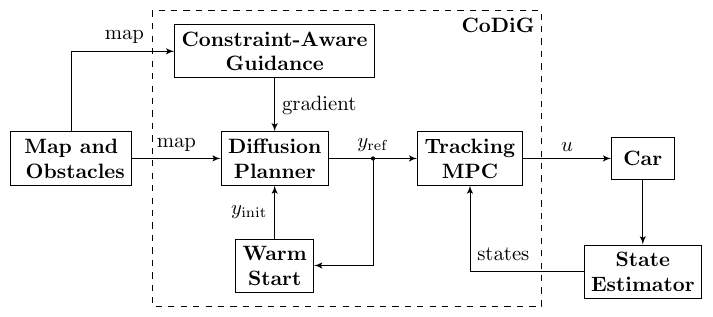} 
  \caption{Flowchart of the proposed CoDiG framework for real-time obstacle avoidance in autonomous racing. The framework integrates a diffusion-based trajectory planner, a constraint-aware guidance module that guides the denoising process, a warm start strategy to accelerate sampling, and a tracking MPC controller. All modules operate within the experimental platform described in Fig.~\ref{fig:platform}.}
  \label{fig:structure}
\end{figure}

The flowchart illustrating how the CoDiG framework enables real-time obstacle avoidance for autonomous racing on the experimental platform is shown in Fig.~\ref{fig:structure}. The core component of the CoDiG framework is a trained diffusion planner module, which generates a safe reference trajectory $y_{\text{ref}}$ capable of avoiding all obstacles. This is achieved by incorporating map and obstacle information, and guiding the sampling process via gradients provided by a constraint-aware guidance mechanism.

To improve sampling efficiency, the reference trajectory generated at the current time point is further used to construct the initial input $y_{\text{init}}$ for the diffusion process at the next time step, via a warm start strategy. This replaces the conventional use of standard Gaussian noise as the initial condition, thereby accelerating the trajectory generation process.

Subsequently, a tracking MPC module computes the control input $u$ required to follow the reference trajectory $y_{\text{ref}}$, based on the current vehicle state estimated by a state estimator module. Finally, the control input $u$ is applied to the vehicle to execute real-time obstacle avoidance.

\end{document}